%% file: paper1.tex
\documentclass[letterpaper]{article} 
\usepackage{aaai24}  
\usepackage{times}  
\usepackage{helvet}  
\usepackage{courier}  
\usepackage[hyphens]{url}  
\usepackage{graphicx} 
\usepackage{natbib}  
\usepackage{caption} 
\usepackage{graphicx}
\usepackage{amsmath}
\usepackage{subcaption}
\usepackage{paralist}
\usepackage{amsthm}
\usepackage{longtable}
\usepackage{multirow}
\usepackage{booktabs}

\usepackage{algorithm}
\usepackage{algorithmic}

\usepackage{newfloat}
\usepackage{listings}
\DeclareCaptionStyle{ruled}{labelfont=normalfont,labelsep=colon,strut=off} 
\floatstyle{ruled}
\newfloat{listing}{tb}{lst}{}
\floatname{listing}{Listing}
\title{Foundations for Unfairness in Anomaly Detection - Case Studies in Facial Imaging Data}

\author {
    Michael Livanos, Ian Davidson
}
\affiliations {
    University of California, Davis\\
    mjlivanos@ucdavis.edu, indavidson@ucdavis.edu
}

\begin{document}



\maketitle

\begin{abstract}
  \input{00abstract}
\end{abstract}


\section{Introduction}
\label{sec:intro}
\input{01intro}

\section{Background and Related Work}
\label{sec:related}
\input{05related}

\section{Four Reasons for Unfairness And Their Measurement}
\label{sec:prelim}
\input{02prelim}

\section{Experimental Results - Who Is AD Unfair To?}

Here we answer the question: Who are the groups of individuals most adversely affected? Following this, we explore more nuanced inquiries, such as whether the unfairness is attributable solely to the data, the algorithm, or a combination of both. In the subsequent section, we aim to investigate the underlying reasons for the unfairness inherent in AD.

Our experiments consist of two core AD algorithms: A reconstruction based autoencoder anomaly detection algorithm (hereby referred to as AE) and Deep one-class SVDD \cite{ruff2018deep}. As mentioned earlier, clustering-based AD is a generalization of one-class algorithms and the AE methods. Our datasets consist of the CelebFaces Attributes Dataset\cite{CelebA} (the 50,000 instance version to reduce compute) which consists primarily of popular individuals in the movies, music, or arts whilst our Labeled Faces in the Wild\cite{LFW} consists of approximately 13,000 instances and includes a wider variety types of popular individuals such as politicians, sports stars, and criminals. Attribution for CelebA is given and attribution for LFW is provided by\cite{LFWAttributes}. These two datasets were chosen as they are \underline{well-annotated}, including analyses of labeling error, and have been extensively studied. Among all of our datasets, we test a total of 63,233 facial images covering 111 attribute tags. We examine each algorithm individually for a total of 222 data points on fairness. Both the CelebA and LFW data sets are publicly available.

For each dataset and algorithm, we determine the unfairness of each group using the Anomaly DIR. Results are collected over five random-initializations of the network and the median results for each property are reported. The list of all raw results is in the appendix, below we outline some key insights.

\begin{figure}[ht]
    \centering
    \includegraphics[width=\linewidth]{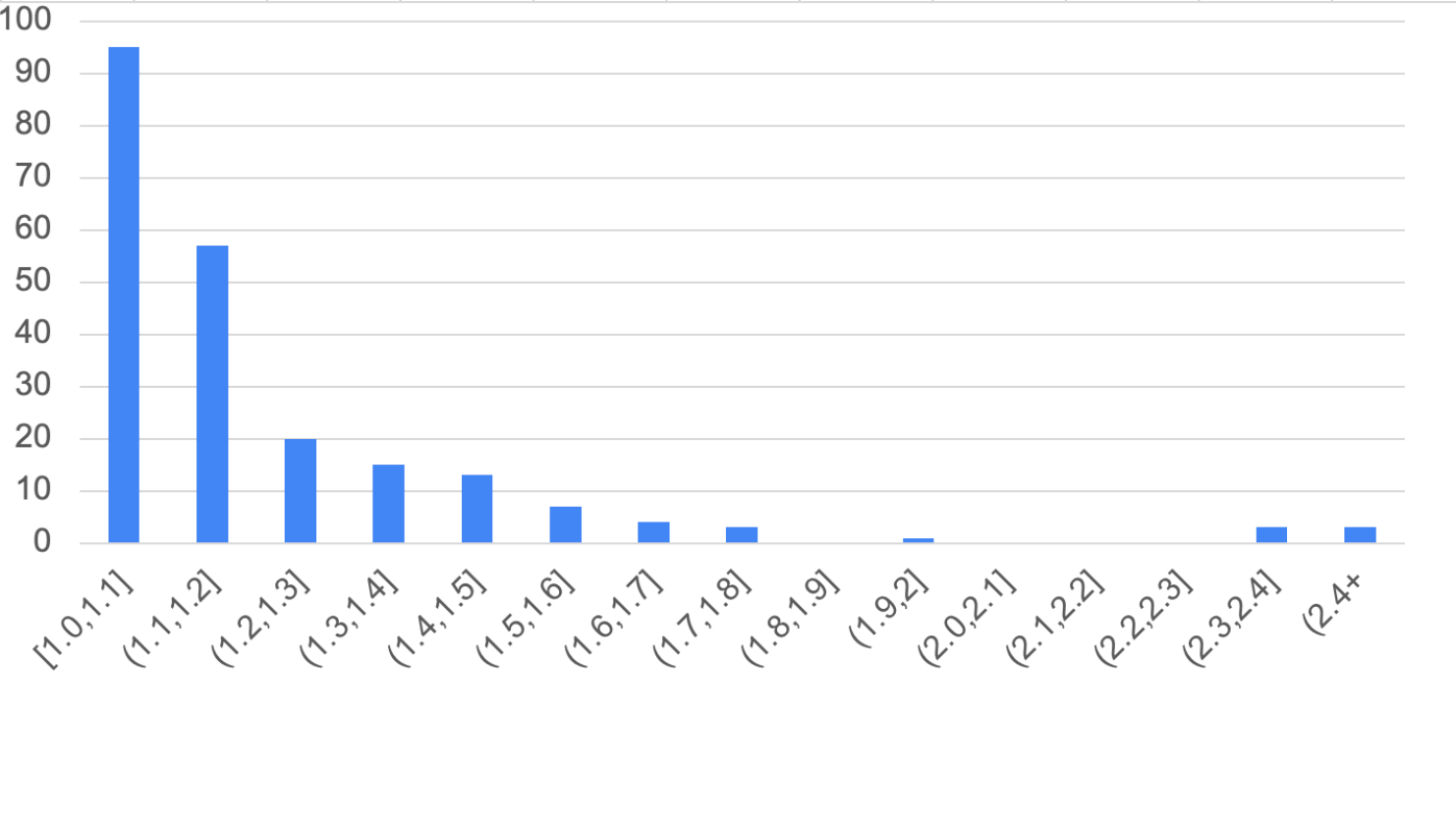}
    \caption{A frequency distribution of the Anomaly DIR score versus how often it occurs across all algorithms and datasets.}
    \label{fig:histogram}
\end{figure}

\smallskip
\noindent
\textbf{The Algorithms are Overwhelming Fair to Most Groups.} In total amongst both the two algorithms and two datasets there are 222 groups and a frequency distribution shows that overwhelmingly the algorithms are fair with respect to over 70\% of the groups as shown in Figure \ref{fig:histogram}. A score of less than 1.2 indicates that the occurrence of the group in the anomalies is not more than 20\% greater than the rate of all other groups (together) being labeled anomalies.

However, there are significant examples of unfairness whose properties we now discuss.

\smallskip
\noindent
\textbf{Few Groups Are \underline{Always} Treated Unfairly.}
We found that there are several groups that are always (regardless of algorithm or dataset)  treated unfairly but they are relatively rare.
These include the groups centered around weight having the annotations \texttt{Chubby, Double-Chin} and those centered around very unusual image properties such as \texttt{Wearing-Hats}. This is not unexpected given a very rare group with unusual properties (not shared by other groups) are unlikely to be well compressed.
In total less than 2\% of all groups are treated unfairly all the time.

\smallskip
\noindent
\textbf{Unfairness Varies  Due to Both Algorithm and Dataset.}
A more likely occurrence is that some groups are treated very unfairly but only for some datasets and some algorithms. Table \ref{tab:crosstab} shows in bold groups treated unfairly  (the Anomaly DIR is shown in parentheses) but \underline{only} for that dataset and algorithm combination. For other algorithm-dataset combinations, they are treated fairly as the Table shows. This result is surprising and shows the strong interaction between the algorithm and the data. Consider that the AE method labeled $\neg \texttt{No Beard}$ (reported as "\texttt{Beard}") in the CelebA dataset at a rate over 3 times greater than the other groups. Yet, the SVDD algorithm on the very same dataset produced just a 1.27 DIR for the \texttt{Beard} group, and in the LFW dataset both algorithms the DIR was below 1.2.

\begin{table}[]
    \centering
    \begin{tabular}{l|l|l}
\textbf{}     & \textbf{CelebA}                                                                                                          & \textbf{LFW}                                                                                                            \\ \hline
\textbf{AE}   & \begin{tabular}[c]{@{}l@{}}\textbf{Beard (3.244)}\\ Senior (N/A)\\ Gray Hair (1.053)\\ Unattractive (1.075)\end{tabular} & \begin{tabular}[c]{@{}l@{}}Beard (1.061)\\ \textbf{Senior (1.8)}\\ Gray Hair (1.028)\\ Unattractive Man (1.158)\end{tabular}     \\ \hline
\textbf{SVDD} & \begin{tabular}[c]{@{}l@{}}Beard (1.267)\\ Senior (N/A)\\ \textbf{Gray Hair (2.449)}\\ Unattractive (1.094)\end{tabular} & \begin{tabular}[c]{@{}l@{}}Beard (1.0876)\\ Senior (1.0018)\\ Gray Hair (1.197)\\ \textbf{Unattractive Man (1.566)}\end{tabular} \\ \hline
\end{tabular}
    \caption{Examples of groups treated unfairly only for a particular algorithm and dataset interaction. The Fairness DIR is reported in parentheses and indicates the relative over-abundance of the group in the anomalies. The tag being treated unfairly in these cases is in bold. For example, people with a Beard are 3.224 times more likely to be an anomaly than a normal instance for the AE algorithm applied to the CelebA dataset, though people with beards are treated relatively fairly otherwise. Not that "Senior" is not a tag in CelebA and is therefore absent from the in these cells.}
    \label{tab:crosstab}
\end{table}

\smallskip
\noindent
\textbf{The More Focused The Dataset The More Likely Unfairness Can Occur.}
When we aggregated all fairness DIR scores (see Appendix) for each group and all algorithms we found that the CelebA dataset (Mean DIR = 1.4) causes significantly more unfairness than the LFW dataset (Mean DIR = 1.13).

This is likely due to the CelebA dataset having a much more focused selection bias as it is limited to people who are overwhelmingly in the arts (film, television, music) whereas the LFW dataset consists of a larger representation of popular people. Hence, the definition of normality learned is very specific and there are many ways to deviate from the norm.
Examples of groups that are found to be unfairly treated in the CelebA dataset but NOT the LFW dataset are: \texttt{Wearing Hat, Big Nose, Eye-Glasses, Goatee, Wavy-Hair}.

\smallskip
\noindent
\textbf{The More Focused The Algorithm The More Likely Unfairness Can Occur.}

\begin{figure}
    \centering
    \includegraphics[width=\linewidth]{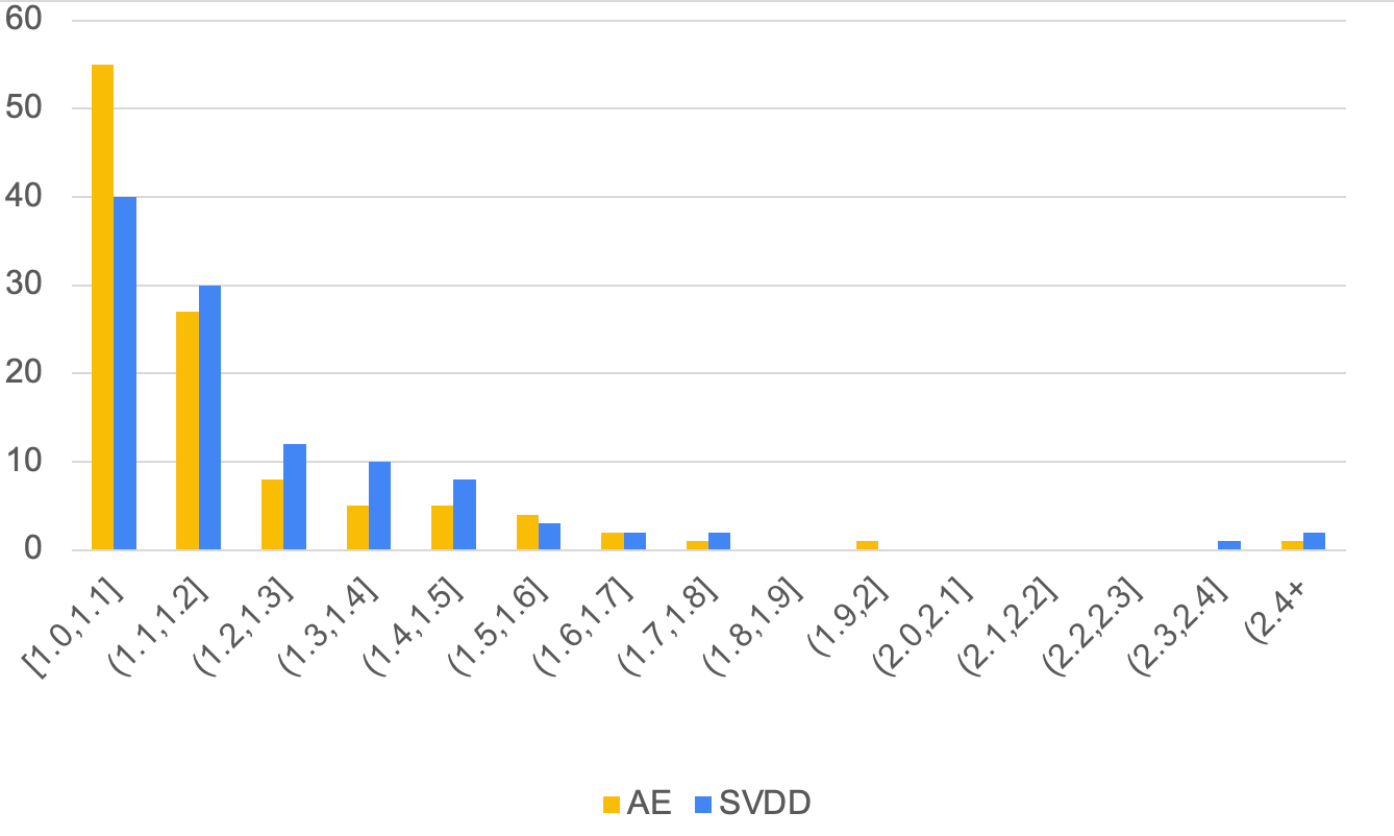}
    \caption{A frequency distribution of the Anomaly DIR score by algorithm. We see that the AE with a more flexible definition of normality is more fair.}
    \label{fig:histAEvsSVDD}
\end{figure}

Similarly, the way the algorithm defines normality is influential in who it identifies as an anomaly. The SVDD algorithm has the strictest definition of normality as it attempts to find just one group of normal instances (centered around $c$ see equation \ref{eq:svdd}) whereas the AE algorithm with $k$ encoding nodes can in practice (assuming perfect disentanglement) find at least $k$ definitions of normality. Hence not surprisingly the SVDD algorithm is more unfairer than the AE algorithm as shown by the histogram of unfairness for both algorithms in Figure \ref{fig:histAEvsSVDD}. 

\section{Experimental Results - Why is AD Unfair}
\label{sec:empirical}

Here we attempt to experimentally answer the following questions:
\begin{itemize}
    \item How strong are our four properties correlated to unfairness?
    \item How are our four properties related to each other and in particular is there a hierarchical structure to them?
    \item How can these properties be combined to create a model to explain unfairness in anomaly detection?
\end{itemize}

\subsection{Relationship between Unfairness and Each Property}

\begin{figure*}
    \centering
    \begin{subfigure}[b]{0.49\textwidth}
        \centering
        \includegraphics[width=\linewidth]{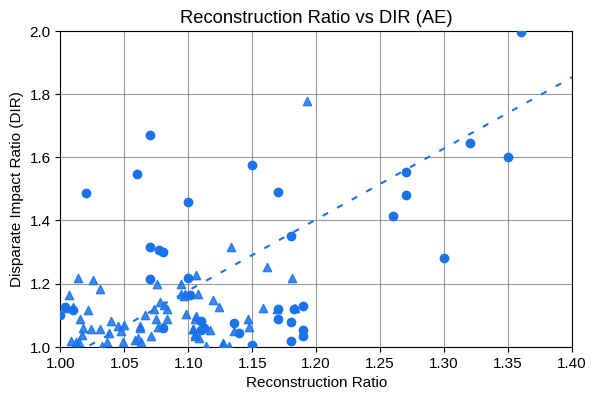}
        \caption{Corr: 0.568, RSQ: 0.334}
        \label{fig:subfig1}
    \end{subfigure}\hfill
    \begin{subfigure}[b]{0.49\textwidth}
        \centering
        \includegraphics[width=\linewidth]{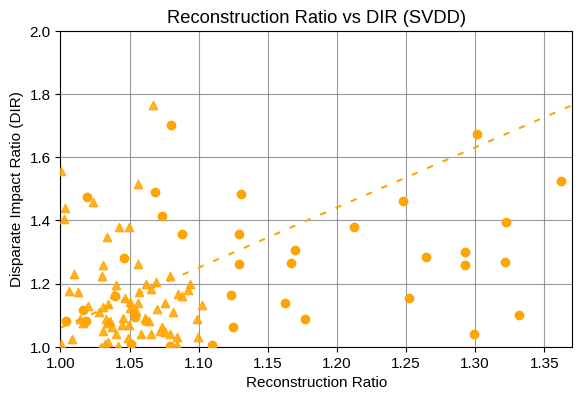}
        \caption{Corr: 0.523, RSQ:0.273}
        \label{fig:subfig2}
    \end{subfigure}
    
    \begin{subfigure}[b]{0.49\textwidth}
        \centering
        \includegraphics[width=\linewidth]{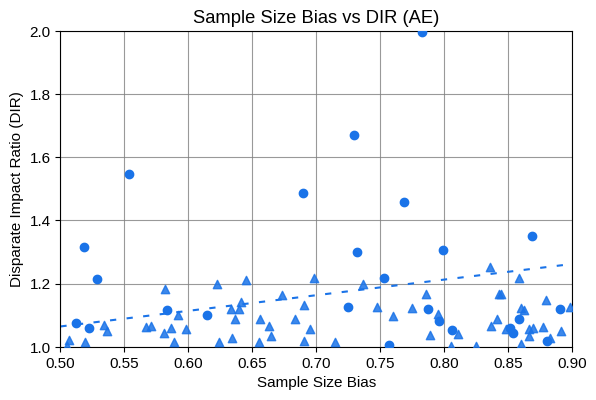}
        \caption{Corr: 0.220, RSQ:0.114}
        \label{fig:subfig3}
    \end{subfigure}\hfill
    \begin{subfigure}[b]{0.49\textwidth}
        \centering
        \includegraphics[width=\linewidth]{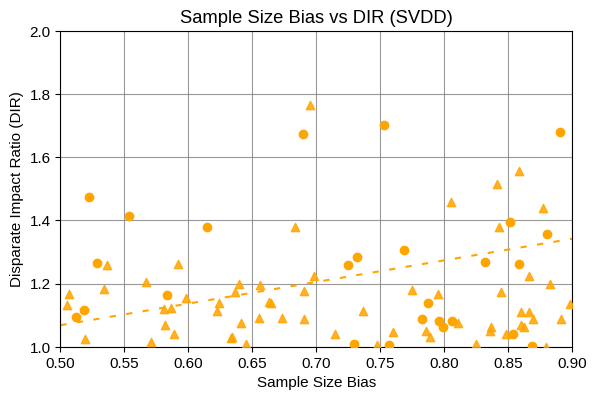}
        \caption{Corr: 0.251, RSQ:0.128}
        \label{fig:subfig4}
    \end{subfigure}
    
    \begin{subfigure}[b]{0.49\textwidth}
        \centering
        \includegraphics[width=\linewidth]{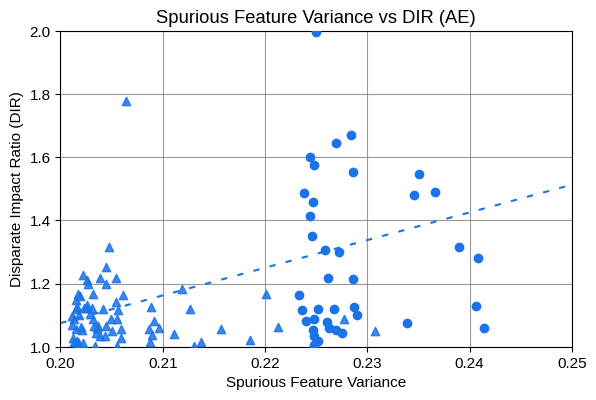}
        \caption{Corr: 0.337, RSQ:0.148}
        \label{fig:subfig5}
    \end{subfigure}\hfill
    \begin{subfigure}[b]{0.49\textwidth}
        \centering
        \includegraphics[width=\linewidth]{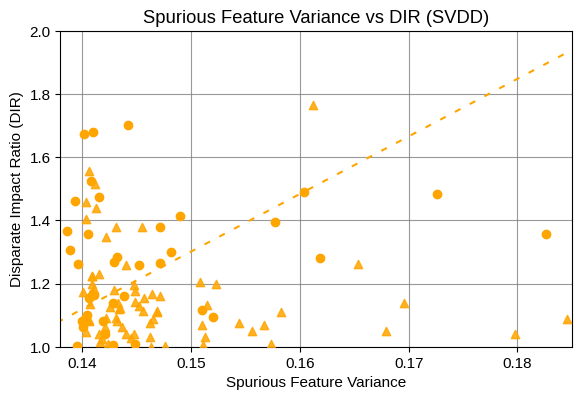}
        \caption{Corr: 0.473, RSQ:0.224}
        \label{fig:subfig6}
    \end{subfigure}
     \caption{(Figure continues on next page)}.
\end{figure*}

\begin{figure*}
    \ContinuedFloat
    \centering
    \begin{subfigure}[b]{0.49\textwidth}
        \centering
        \includegraphics[width=\linewidth]{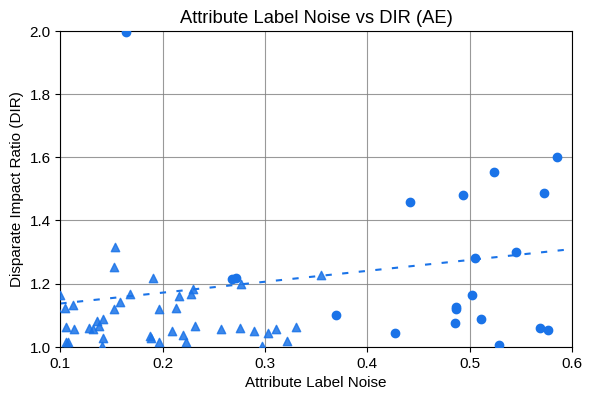}
        \caption{Corr: 0.261, RSQ:0.167}
        \label{fig:subfig7}
    \end{subfigure}\hfill
    \begin{subfigure}[b]{0.49\textwidth}
        \centering
        \includegraphics[width=\linewidth]{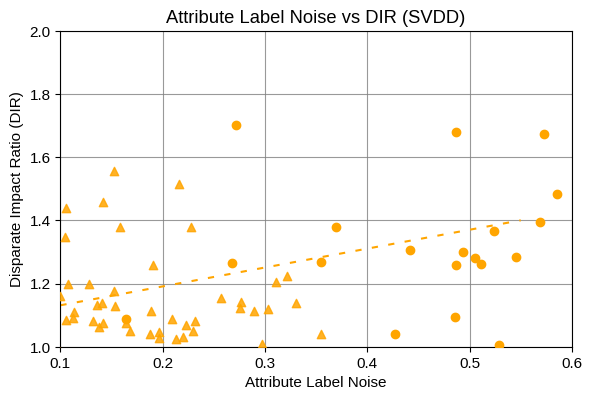}
        \caption{Corr: 0.328, RSQ:0.108}
        \label{fig:subfig8}
    \end{subfigure}
    \caption{Plot of different properties against their DIR (unfairness) with the larger the value the more of the property/unfairness. Trendlines are created by minimizing $R^2$ values. Each mark represents one group. Color denotes algorithm (blue for the AE anomaly detector and orange for the SVDD anomaly detector) and mark denotes dataset (circle for CelebA, triangle for LFW).}
    \label{fig:propertyRelationshipGraphs}
\end{figure*}

Our experiments (see Figure \ref{fig:propertyRelationshipGraphs}) demonstrate strong (Pearson) correlations and moderate to strong RSQ (R-squared values of the regression trendline) for each of the properties studied. Each plot shows the results for two datasets (CelebA and LFW) with each data point representing a group of individuals. A positive trend line indicates positive Pearson correlation (see sub-titles of plots for exact values) and we see that incompressability is the most strongest property correlated with unfairness, then Spurious features, then Attribute label noise, and finally Sample Size Bias. This is an interesting result as earlier seminal results showed that AD using facial images \cite{zhang2021towards} was unfair due to an under-representation of African Americans and Males in the underlying datasets. 

However, it is also clear that no individual property explains unfairness completely by itself. 
This is shown as each graph has points that not only do not fit the trendline, but are contradictory to the relationship implied by the overall data. Further investigation (see next subsection) reveals that when one property fails to explain why that attribute is anomalous, another one typically will. 

For example, the group \texttt{Bags Under Eyes} (from CelebA) under the AE model has a reconstruction ratio of only 1.077 (it is easy to compress), but a DIR of 1.31 (it is treated unfairly). Following the trend, the expected reconstruction ratio at a group with this DIR would be approximately 1.17. Further, this group has only 20.1\% representation, though looking at the DIR one would expect only half that. This group's treatment, however, is explained by the spurious feature variance, as it sits nearly perfectly on the trendline. Similarly, the group \texttt{Gray Hair} (from LFW) under Deep SVDD was towards the far end of spurious feature variance at 0.180, but has extremely low anomaly DIR score at 1.04 (i.e. was treated fairly), though it sits just above the trendline for attribute label noise at 1.05.

A full list of these attributes and their squared error for all trendlines is available in the  Appendix, and one can see that every tag can be explained by at least one of these properties with high fidelity, with the average sum of square errors being only 0.00351 (std 0.006498), supporting our claim that unfairness in anomaly detection setting can be typically explained by one of these four properties. This claim is rigorously tested in Section \ref{sec:Hypothesis}.

\subsection{Relationship between Multiple Properties}

\begin{figure}[]
    \centering
    \includegraphics[width=0.8\linewidth]{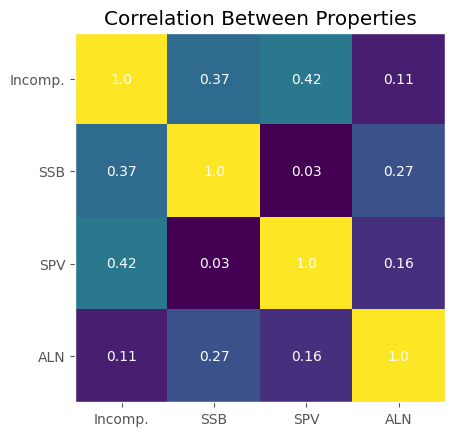}
    \caption{Correlation matrix for all four properties of the model. Pearson correlation is written in each box and is consistent with color (yellow is large, purple is small).}
    \label{fig:correlationMatrix}
\end{figure}

We also examine the correlation between the different properties. This analysis is useful in examining potential redundancies and creating our model of unfairness for anomaly detection. Figure \ref{fig:correlationMatrix} examines these relationships. Some features are, indeed, positively correlated with each other, though none have high enough correlation to suggest that they are redundant with each other. In the subsequent subsection, we examine this claim more rigorously via a hypothesis test.

\subsubsection{Hypothesis Testing of Relationship Claims}
\label{sec:Hypothesis}

In order to test our claims, we create four hypotheses that we verify through hypothesis significance-testing. Those are:

\begin{itemize}
    \item H1: No individual property is sufficient to always explain unfairness.
    \item H2: The properties, when combined into a multiple regression, are sufficient to explain unfairness.
    \item H3: No properties of the multiple regression are redundant and all are needed.
    \item H4: The results of H2 are significant in that when one property fails to predict unfairness, another does.
\end{itemize}

\begin{figure}[h]
    \centering
    \begin{subfigure}[t]{0.45\textwidth}
        \centering
        \includegraphics[width=\textwidth]{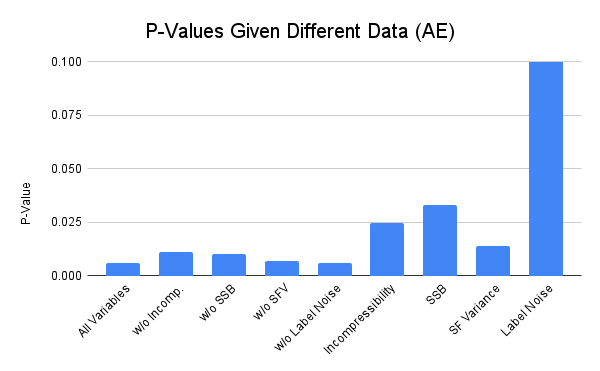}
        \label{fig:hypothesisAE}
    \end{subfigure}
    \hfill
    \begin{subfigure}[t]{0.45\textwidth}
        \centering
        \includegraphics[width=\textwidth]{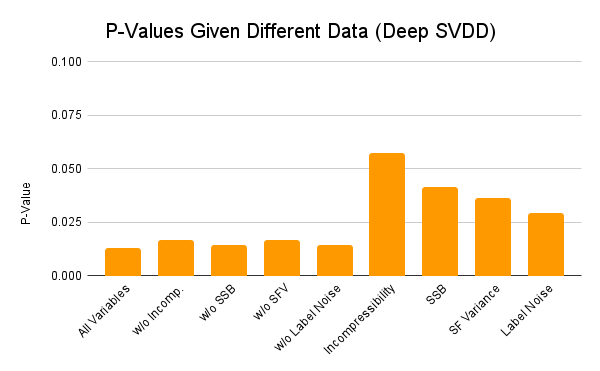}
        \label{fig:hypothesisSVDD}
    \end{subfigure}
    \caption{P-Values for the hypotheses H1-H3. The leftmost bar demonstrates that, when all properties are considered, unfairness can be predicted with a very high degree of precision, rejecting the null hypothesis $\text{H2}_0$. The next three rows demonstrate that the model is not as powerful if one property was left out, rejecting the null hypothesis $\text{H3}_0$. Finally, the higher P-values for the simple regressors indicate that no single feature can be used as a model of unfairness, rejecting null hypothesis $\text{H1}_0$.}
    \label{fig:hypothesis}
\end{figure}

Null hypothesised $\text{H1}_0-\text{H4}_0$ are constructed straightforwardly. To create the significance test for H1, we perform an F-test on individual regression models crafted from the relationship between each property and DIR. The results of this F-Test (visualized in Figure \ref{fig:hypothesis}) indicate that individual properties are reasonable though comparably weak predictors of unfairness, with P-values ranging from 0.0137-0.0986 for the AE model and 0.0279-0.0571 for Deep SVDD. Therefore, we reject the null hypothesis $\text{H1}_0$ and validate hypothesis H1.

To test hypotheses H2 and H3, we construct a multiple-regression model. Specifically, this is a \underline{stacked} multiple regression where the meta-function selects the best individual model for the datum. To validate H2, we create such a multiple-regression using all four of the properties (the "full" model). This yields P-Values of 0.00589 for the AE model and 0.0127 for Deep SVDD, significantly lower than those of the respective single-regression models, and indicating that using all four properties is sufficient to explain how unfairness occurs. We reject the null hypothesis $\text{H2}_0$ and validate hypothesis H2.

For H3, we conduct a similar experiment except we leave one property out. In every case, the resulting multiple regression models were worse than the full model, with P-Values ranging from 0.00674-0.0109 for the AE model and 0.0138-0.0164 for Deep SVDD, all greater than that of the full model, indicating that every property is necessary and none are redundant. We reject the null hypothesis $\text{H3}_0$ and validate hypothesis H3.

One may object to the multiple-regression models used above, given that the model as described will monotonically increase in predictive power given more properties. It is important to note that this model matches the central claim of this paper - that unfairness with respect to a group occurs because of one of the four properties described, though one may still be wary of the statistical significance of the reported results given the technique. To resolve these concerns, we demonstrate that our model is not just combining the predictive power of four different already powerful predictors, but rather when one model fails it is because it is explained by one of the other properties.

To validate this claim, we construct fabricated distributions similar to those of Figure \ref{fig:propertyRelationshipGraphs}. Specifically, unfairness is kept the same, and we create distributions of random fake data which has the same correlation and RSQ as all of those shown. This is accomplished by, for each property, finding random points (sampled across a uniform distribution) along the X-axis, giving them fabricated values perfectly in line with the correlation, and then adding noise such that the correlation is maintained and the RSQ matches that of the actual measured properties. Then, we create the same full model of the multiple regression and measure the P-value. We repeat this process 10,000 times to get 10,000 such distributions.

The distributions therefore should be statistically similar to our real data, but there is no reason to believe that when one of the fabricated models fails, another will explain the unfairness. To validate hypothesis H4, we measure the number of times the fake distributions produce P-values under that of the real data. If the statically similar fabricated data cannot match the predictive performance of our models, this would validate hypothesis H4.

In the case of the AE model, the fabricated data averaged a P-value of 0.0194 with a standard deviation of 0.00629 and never beat the full model's P-value of 0.00589. Similarly, the model simulating Deep SVDD's data yielded an average P-value of 0.0173 with a standard deviation of 0.00304. Out of the 10,000 trials, only 5 yielded lower P-values. Therefore, we reject the null hypothesis $\text{H4}_0$ and validate hypothesis H4. Our model does not simply take four independent good predictors of anomaly and get good statistical results but rather holds the property that when one fails, another property explains it.

\subsection{A Proposed Model Of Unsupervised Unfairness Relationships}

\begin{figure}[htbp]
    \centering
    \includegraphics[width=0.75\linewidth]{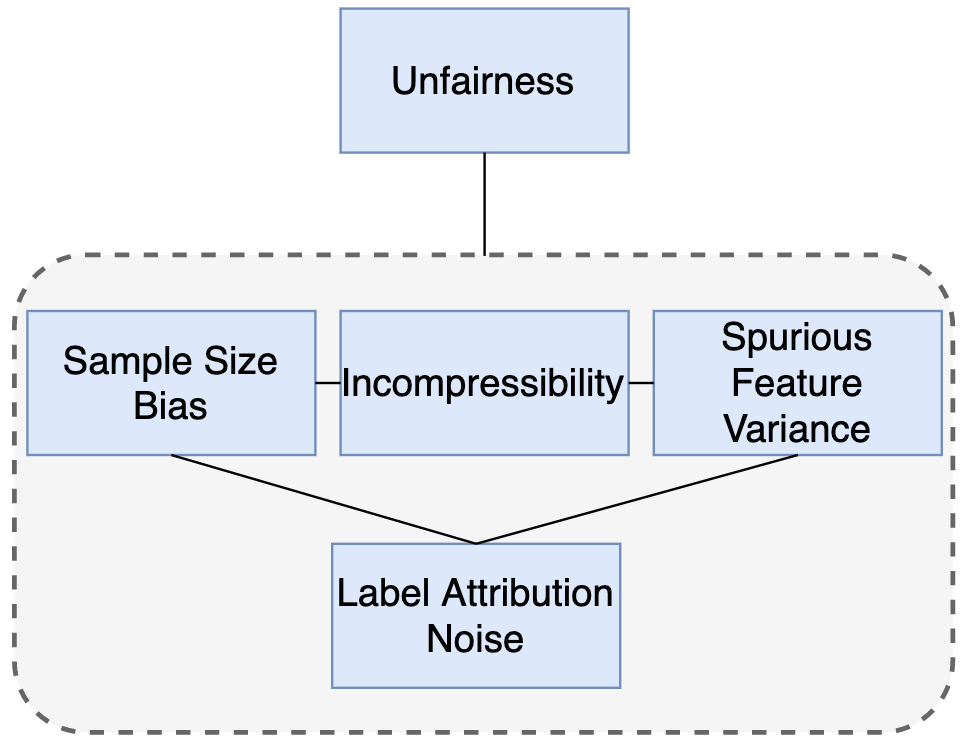}
    \caption{Our model of unfairness determined from our stacked multi-regression model. Compare with the expected model without any analysis in Figure \ref{fig:overview}.}
    \label{fig:trueModel}
\end{figure}

Given the resulting hypothesis tests, we craft our model of unfairness in unsupervised learning. Figure \ref{fig:trueModel} provides a graphical representation of this model. Edges between properties indicate a relationship (binarized to be correlated at $\geq$ 0.15). This is supported by the high correlation between each of these properties and unfairness (Figure \ref{fig:propertyRelationshipGraphs}), the result that the properties together form a uniquely powerful multiple-regression to explain unfairness (H2, H4), that no single feature could do this alone (H1), and that no property is redundant (H3).

\section{Conclusion, Limitations, and Future Work}
\label{sec:conclusion}
\input{06conclusion}


\bibliography{paper1.bib}


\appendix
\section*{Appendix}
\input{appx.tex}

\end{document}

%% file: 00abstract.tex
Deep anomaly detection (AD) is perhaps the most controversial of data analytic tasks as it identifies entities that are then specifically targeted for further investigation or exclusion. Also controversial is the application of AI to facial imaging data. This work explores the intersection of these two areas to understand two core questions: \emph{"Who''} these algorithms are being unfair to and equally important \emph{"Why''}. Recent work has shown that deep AD can be unfair to different groups despite being unsupervised with a recent study showing that for portraits of people: men of color are far more likely to be chosen to be outliers. We study the two main categories of AD algorithms: autoencoder-based and single-class-based which effectively try to compress all the instances with those that can not be easily compressed being deemed to be outliers. We experimentally verify sources of unfairness such as the under-representation of a group (e.g. people of color are relatively rare), spurious group features (e.g. men are often photographed with hats), and group labeling noise (e.g. race is subjective). We conjecture that lack of compressibility is the main foundation and the others cause it but experimental results show otherwise and we present a natural hierarchy amongst them.

%% file: 01intro.tex
Anomaly detection (AD) is a central part of data analytics and perhaps the most controversial given that it is employed for high-impact applications that identify individuals for intervention, policing, and investigation.  Its use is prevalent to identify unusual behavior in finance (transactions)\cite{financial, zamini2019comprehensive}, social media (posting and account creation)\cite{yu2016survey, savage2014anomaly}, and government services (medicare claims)\cite{zhang2017anomaly, bauder2017multivariate}.

Perhaps one of the most controversial applications of AI is to facial imaging. 
This is due to our faces being uniquely identifying and personal. Further, the AI's ability to identify us and make decisions (without consent) crosses many cultural and legal barriers \cite{garvie2016perpetual}.
Existing work on facial data has focused predominantly on facial recognition, that is, given a large collection of people in a known database, identify if any of them occur in an image. Though legislation and progress have been made towards regulating facial recognition technology \cite{almeida2022ethics} other technologies in particular AD involving facial images are starting to emerge which gives rise to new ethical considerations and understanding.

\begin{figure}[!t]
	\centering
	\vskip -0.1in
	\includegraphics[width=0.43\textwidth]{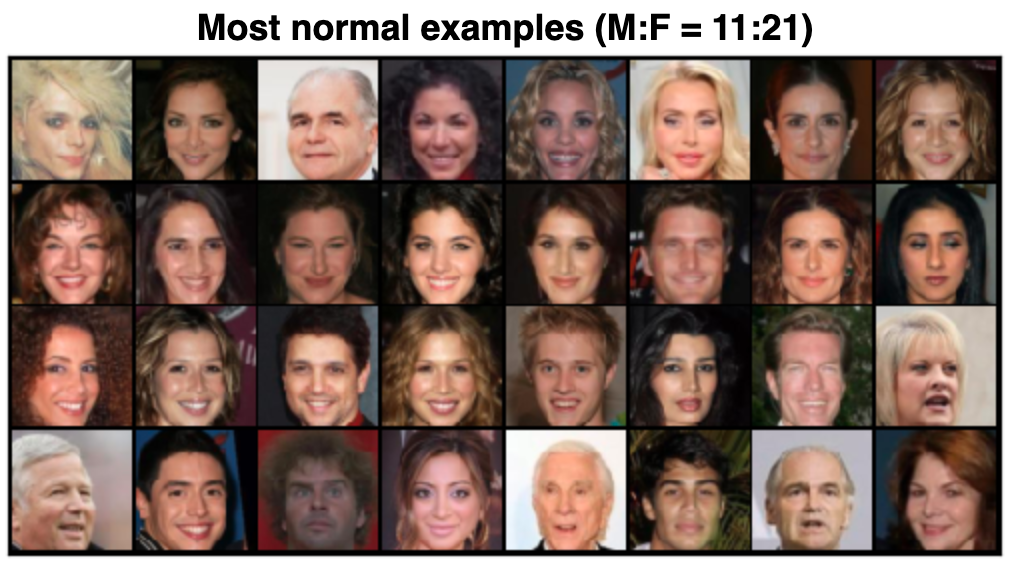}
	\label{1a}
	\hfill
	\includegraphics[width=0.43\textwidth]{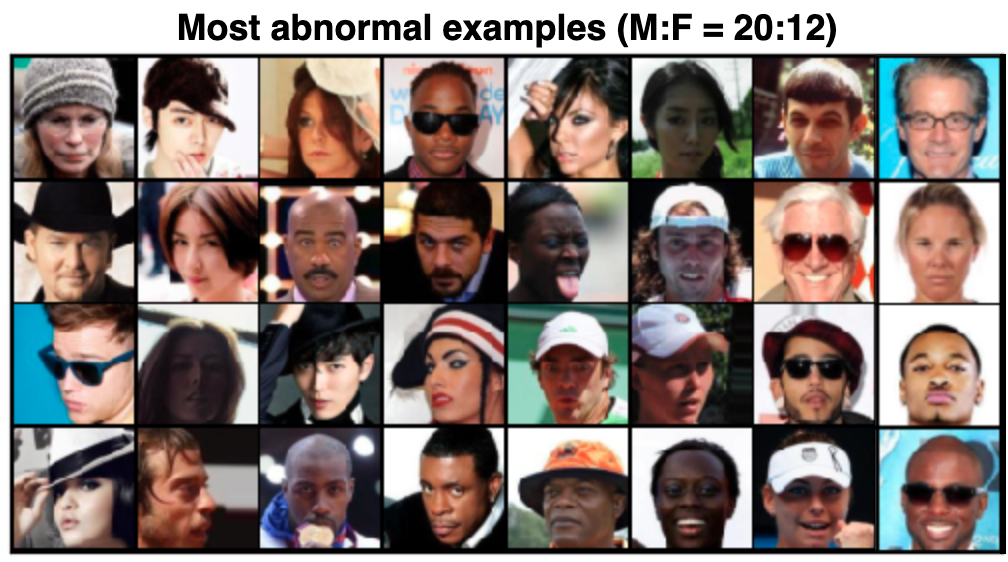}
 \caption{Example of AD Being Unfair When Applied to Facial Imaging Data. Reproduced from \cite{zhang2021towards}.} 
 \label{fig:demo}
\end{figure}

Previous work \cite{zhang2021towards} has just begun to explore the unfairness at the intersection of AD applied to facial imaging data. For example, our previous work showed that applying AD to a collection of celebrity images overwhelmingly showed the anomalies being people of color and males (see Figure \ref{fig:demo}). However, our previous work was mainly focused on making AD algorithms fairer. We recreate our earlier results for not only the one-class AD method and the celebrity image dataset the authors used but also for the popular auto-encoder AD method and a more challenging dataset (Labeled Face In The Wild\cite{LFW}). 

Our experimental section attempts to address the ``Who" and ``Why" questions.
We create a measure of unfairness (Disparate Impact Ratio (DIR)) which measures how over-represented a protected group (or its complement) is in the anomaly set. We then experimentally investigate \underline{who} these algorithms are being unfair to and more nuanced questions such as is the same group always being treated unfairly regardless of algorithm. 
We also explore \underline{why} an unsupervised algorithm can be biased. We conjecture four main foundations of unfairness, propose metrics to measure them, and outline a series of experiments to test a hypothesis on how they are structured. 

The contributions of this work as are as follows:
\begin{itemize}
    \item We study the ``Who'' and ''Why'' questions when anomaly detection is applied to facial imaging data - a topic to our knowledge has not been addressed before.
    \item Our experiments addressing the ``Who" question show that group-level unfairness is due to an interaction between the dataset and the algorithm.
    \item We conjecture four main reasons for the ``Why" question: i) incompressibility, ii) sample size bias (SSB), iii) spurious feature variance (SFV) within a group, and iv) attribute/group labeling noise (ALN).
    \item We postulate an intuitive structure to our conjectured reasons, showing it is not empirically verified, but our experimental results suggest an alternative structure.
\end{itemize}

We begin by discussing background and related work. We then introduce how we measure unfairness in AD and our four proposed foundations of unfairness. Next, our experimental results addressing the ``Who" and ``Why" questions are presented after which we discuss and conclude our work.

%% file: 05related.tex
\textbf{Applications of AD to Facial Data.}
AD algorithms have been used on imaging data for a variety of reasons. Perhaps the most ubiquitous is for data cleaning where anomalies are viewed as being ``noise" \cite{ng2014data} which are removed and then a downstream supervised algorithm is applied. 
However, if the AD algorithm is biased this creates an under-representation in the down-stream training tasks.

Another common use of AD is to view the outliers as ``signal" and in doing so flag the outliers for extra attention. Examples include using AD to identify facial expressions to recognize emotions \cite{zhang2020outlier} such as surprise. However, if the AD is biased towards some groups this will over-predict certain emotions for certain groups. Similarly, AD can be used to identify aggressive behavior \cite{cao2021outlier}. However, if the AD has a bias towards some groups this will incorrectly identify the group as being overly aggressive.

\noindent
\textbf{Source of Bias.} It has been well established that supervised learning algorithms can have bias due to a variety of reasons. In particular class labeling bias has been extensively studied in the context of the Compas dataset \cite{compas}. Even though features (e.g. race) associated with this bias are removed, deep learning offers the ability to learn surrogates (e.g. zip code)\cite{labelBias}. 

The work on fair AD starts in 2020 \cite{davidson2020framework,abraham2021fairlof} and has shown that AD algorithms can cause bias. Most work has focused on how to correct unfairness for a certain algorithm. This involves understanding the limitations in the algorithm's computation and then correcting for it. This has been explored for classic density-based methods such as LOF \cite{abraham2021fairlof} and deep learning methods for autoencoder \cite{shekhar2021fairod}, one class \cite{zhang2021towards} and multi-class deep AD methods. However, despite this earlier body of work, there has been surprisingly little work discussing what produces unfairness in unsupervised anomaly detection.

%% file: 02prelim.tex
Here we outline our four premises for unfairness in AD and explain them at a conceptual level using Figure \ref{fig:demo}.  We then describe how we measure them.

\subsection{Incompressability of Data}
We begin by discussing how AD methods work in particular what causes an instance to be an outlier. Deep AD methods at their core employ compression either directly or indirectly. Instances that cannot be compressed well are deemed outliers and if a group is unusual in some sense it will be unfairly treated as it will be hard to compress and hence overwhelmingly flagged as an outlier.

To understand this further, we present a common taxonomy of anomaly detection algorithms\cite{pang2021deep}.

\noindent
\textbf{Autoencoder for Anomaly Detection.}
Let $\phi_{e}$ be the encoding network which maps the data $X$ into the compressed latent space and $\phi_{d}$ be the decoding network which maps the latent representation $\phi_{e}(X)$ back to the original feature space\cite{Hinton1989}. 
Given the network parameters $\theta_{e},\theta_{d}$ the standard reconstruction objective to train the autoencoder is:
\begin{equation}
\label{eq:ae}
     argmin_{\theta} \left( \frac{1}{n} \sum_{i=1}^{n} \|x_i - \phi_{\theta_d}(\phi_{\theta_e}(x_i))\|^2 + R \right)
\end{equation}
The term $R$ denotes the regularization to the encoder and decoder. The anomaly score $s(x)$ for instance $x$ is calculated from the reconstruction error:
\begin{equation}
\label{eq:outlier_score_ae}
    s(x) = {||x - \phi_{\theta_d}(\phi_{\theta_e} (x) )||}^2
\end{equation}
Here clearly an outlier is defined as being an instance that the AE cannot easily compress and hence cannot easily reconstruct\cite{Japkowicz1995}.

\noindent
\textbf{One-Class/Cluster Anomaly Detection}
Next, consider one class anomaly detection which is still unsupervised. Given the training data of instances $X \in {R}^{n \times d}$, one class AD method such as the the popular deep SVDD \cite{ruff2018deep} network is trained to map all the $n$ instances close to a fixed center $c$. 
Denote function $\phi$ as a neural network with parameters $\theta$ the training objective function is:

\begin{equation}
\label{eq:svdd}
    argmin_{\theta} \frac{1}{n} \sum_{i=1}^{n}{||\phi_{\theta}({x}_i) - c||}^2 + R
\end{equation}

where the term $R$ represents the regularization function. 
Then the anomaly score is naturally the distance to $c$.

\begin{equation}
    s(x) = {||\phi_{\theta}(x) - c||}^2
    \label{eq:svdd_outlier_score}
\end{equation}

Here the aim is to compress all points to map onto a central point $c$ and those that cannot be are deemed outliers.

\noindent
\textbf{Deep Clustering for Anomaly Detection}
Deep Embedded Clustering (DEC) \cite{xie2016unsupervised} is one of the earlier deep clustering methods that combines representation learning with clustering using a clever self-supervision approach.  Recently this work was extended to perform outlier detection \cite{song2021deep}.

The distance a point is from its closest centroid $\{c_1, ... c_K\}$ is naturally an anomaly score $s(x)$: 
\begin{equation}
    s(x) = \frac{\min_{k \in [1,K]}||\phi_{\theta_e}(x)-c_k||^2}{\max_{j \in [1,n]  \land  m_j = k}||\phi_{\theta_e}(x_j)-c_k||^2}
    \label{eq:outlier_score_deepclustering}
\end{equation}
where $m_j = k$ denotes instance $x_j$ belongs to cluster $c_k$, $K$ denotes the total number of clusters, and $\phi_{\theta_e}(x_i)$ is the deep learner embedding function.

The core idea here is an extension to the one-class AD method mentioned earlier but extended to $k$ clusters.

\subsection{Causes Beyond Incompressibility} 
\begin{figure}
    \centering
    \includegraphics[width=0.45\textwidth]{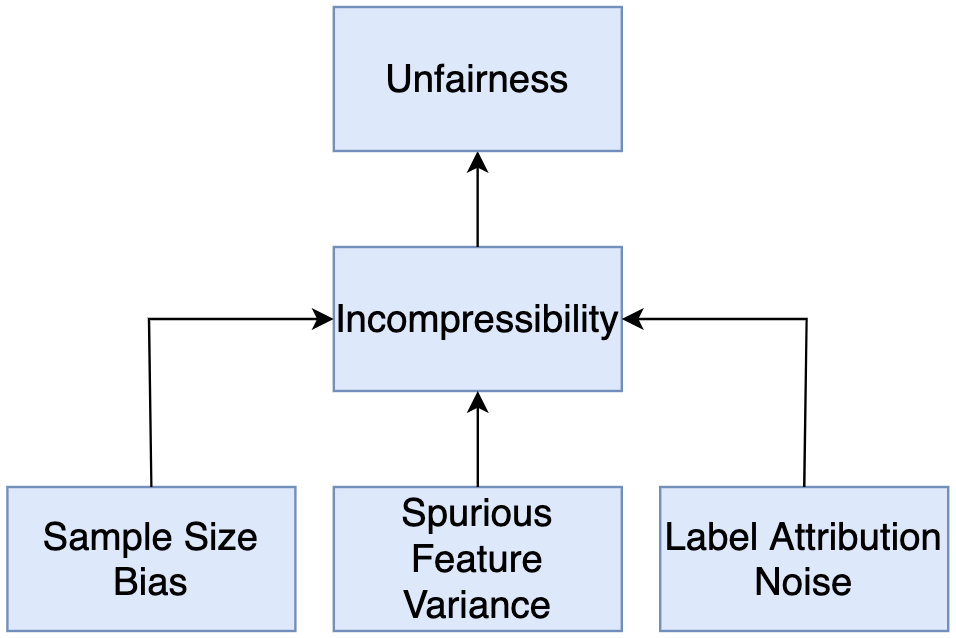}
    \caption{A Diagrammatic view of the  \underline{expected} reasons behind biased outlier detection.}
    \label{fig:overview}
\end{figure}

The above states that outliers are inherently points that the deep learner cannot compress. Hence it is natural to consider reasons why a deep learner cannot compress a group as being a key issue for unfairness. Here we \underline{conjecture} three main reasons with the view they are related to biased outliers as shown in Figure \ref{fig:overview}.

\textbf{Group Underrepresentation.}
Here we have a group that is relatively rare in the dataset but has some unique properties so the deep learner cannot compress it well. For example in Figure \ref{fig:demo} many outliers are African Americans as they only consist of under 15\% of the dataset hence the deep learner uses its limited encoding space to encode more populous properties.

\textbf{Spurious Features for Groups.}
In this situation, the group has a property that is not critical for the outlier detection task but is highly variable. For example in Figure \ref{fig:demo} many groups who are over-represented in the outliers wear different styles of hats.

\textbf{Label Attribution Noise.}
Here the labeling of a group is inaccurate and hence can be a reason a group is labeled as being overly abundant in the outlier group. For example in Figure \ref{fig:demo} the second to the bottom line of outliers all have the tag \texttt{Male} but this is erroneous.

\subsection{Measurements of Unfairness and Four Properties}

Before discussing our empirical results, we first define each of the properties and how anomaly unfairness is measured. Many of these metrics are the maximum between some expression and their reciprocal. This is because the presence of a tag is equally important as the absence of a tag: for example, disparate treatment of young people and disparate treatment of old (i.e. not young) people are equally important phenomena to study. We first describe how we measure unfairness for anomalies and then how we measure our four properties.

\noindent
\textbf{Anomaly DIR:} The unfairness of an AD algorithm's output for particular group $a$ is measured by the disparate impact ratio (DIR), which is \cite{feldman2015certifying}:

\begin{equation}
\label{eq:DIR}
\begin{split}
    DIR(X,AD,a) = \max\Biggl(&\frac{P(AD(X)=1|A=a)}{P(AD(X)=1|A=\neg a)}, \\
    &\frac{P(AD(X)=1|A=\neg a)}{P(AD(X)=1|A= a)}\Biggr)
\end{split}
\end{equation}

Here $X$ is the dataset the AD algorithm ($AD$) has made predictions (normal vs anomaly) with $AD(x)=1$ implying $x$ is an anomaly and $AD(x)=0$ implying it is a normal instance, and $a$ is the group in question. This is a natural choice for anomaly detection as it compares the rate at which different attributes are flagged as anomalies, normalized by how often the rest of the data is considered anomalous. It is also the most widely used metric in fair unsupervised learning\cite{fairnessDefinition}. The range for this metric is $[1,\infty)$ with the larger the number the more unfairly group $a$ is treated.

\noindent
\textbf{Incompressibility:} To measure this feature, we extend the typical measure of reconstruction error into the novel metric of reconstruction ratio, which is defined:

\begin{equation}
\label{eq:RR}
\begin{split}
    RR(X,f,a) = &\max\Biggl(\frac{Loss_{MSE}(X,f(X)|A=a)}{Loss_{MSE}(X,f(X)|A=\neg a)}, \\
    &\frac{Loss_{MSE}(X,f(X)|A=\neg a)}{Loss_{MSE}(X,f(X)|A=a)}\Biggr)
\end{split}
\end{equation}

Here $X$ and $a$ are the data used for AD and group again, with $f$ being the autoencoder model (both encoder and decoder). The range of Equation \ref{eq:RR} is therefore also $[1,\infty)$, where a higher number indicates that a group is harder to compress than the rest of the data. For example, a RR of 2 indicates that the attribute/group (or absence of the attribute/group) is twice as difficult to compress than the rest of the data.

\noindent
\textbf{Sample Size Bias (SSB):} SSB (sometimes referred to as representation bias) is determined by the proportion of that tag or lack in the dataset $X$ and is measured as\cite{Suresh_2021}:
\begin{equation}
\label{eq:SSB}
    SSB(X,a) = max(P(A=a|X),P(A = \neg a|X))
\end{equation}

Where $X$ and $a$ are again the data and the group in question. Because all groups are binary (or encoded as one-hot encoding), the range of this metric is $[0.5,1]$, with 0.5 indicating perfect balance of the group (i.e. males and females are equally likely)  and 1 indicating that the group is always on or always off. Most groups will fall between these two extremes.

\noindent
\textbf{Spurious Feature Variance (SFV):} SFV refers to the amount of variance in the background objects in the image and is measured as a proportion of the reconstruction error of the image:

\begin{equation}
\label{eq:SPV}
\begin{split}
    SFV(X,f,a,b)&=1-\max\Biggl(\frac{Loss_{MSE}(X[b],f(X)[b]|A=a)}{Loss_{MSE}(X,f(X)|A=a)}, \\
    &\frac{Loss_{MSE}(X[b],f(X)[b]|A=\neg a)}{Loss_{MSE}(X,f(X)|A=\neg a)}\Biggr)
\end{split}
\end{equation}

Where $X$ is the data, $f$ is the autoencoder, $a$ the tag, and $b$ is a bounding rectangle around the foreground/focus of the image (i.e. the face), either provided by the data or estimated\cite{LFWAttributes}. As the denominator is clearly always greater than or equal to the numerator, SFV ranges between $[0,1]$, where higher values indicate that more error comes from spurious features.

\noindent
\textbf{Label Attribute Noise (LAN):} This is a metric of how noisy the labeling of a particular group is, as provided by the academic literature(\cite{CelebANoise} for CelebA and \cite{LFWAttributes} for LFW). Some groups such as Gender tend to have very low LAN, whereas other tags have very high LAN such as Blurry\cite{LFWAttributes}. We define LAN as:

\begin{equation}
\label{eq:ALN}
    ALN(X,a,a^*) = 1- (P(a=a^*|X)+P(\neg a=\neg a^*|X))
\end{equation}

Where $X$ is the data, $a$ the group in question, and $a^*$ the true label for the group. This property has a range $[0,1]$ where the higher the value the less reliable the group labeling.

%% file: 06conclusion.tex
We study the intersection of the controversial deep AD algorithm with facial imaging data to address the ``Who" and ``Why" questions. We found that overwhelmingly both auto-encoder and one-class deep AD algorithms are fair to most groups. However, due to the compression-based focus, they are unfair to some sub-groups.

With regard to the ``Who" question we found that it was rare to be consistently unfair to the one group and instead unfairness was due to the interaction of the data and the algorithm. In particular, the more focused the dataset and algorithm the more unfairness was found.   

Our study of the ``Why" question aimed at developing a deeper understanding on the effect of data related factors on the fairness as well as detection performance of OD algorithms.
We postulated four hypotheses and found all to be statistically significant by rejecting the null hypothesis. The first hypothesis is that no single property alone is sufficient to explain unfairness. The second hypothesis is when combined the properties can explain unfairness. The third hypothesis is that all properties are relevant and none are redundant and finally, the fourth hypothesis is that the combination of properties is meaningful beyond the predictive power of each individual property.

\noindent
{\bf Limitations.}
The use of groups may have varying degrees of applicability to real-world fairness scenarios. For example, some groups such as \texttt{Male, Black} and \texttt{Young} correspond to legally recognized protected classes \cite{civilrightsact1964, adea1967}, while others such as \texttt{Goatee, Wearing Hat} and \texttt{attractive} may not. However, we believe that this study still provides meaningful insights into the mechanism of unfairness with respect to different people. Real-world protected attributes may be of varying degrees of visibility, as do our groups, and our analysis reflects this.

\noindent
{\bf Future work.} 
Remediation strategies to improve fairness are left out of scope of our investigation.
We briefly discuss them here.
Fairness interventions are typically grouped into three: pre-, post-, and in-processing  strategies, which respectively, modify the input data, modify the output scores or decisions, and account for fairness during model training. 

As we showed, AD unfairness can stem from algorithmic bias alone in the face of natural heterogeneities in the data among or within groups. When this is the case, pre-processing strategies become voided as it is not clear how to modify organic, unbiased data.
Post-processing could select different thresholds for each group separately, as in \cite{corbett2017algorithmic,menon2018cost}, where the group-specific thresholds could either be ``natural'' cut-off values, or selected to optimize demographic parity if it is a desired fairness metric. Note that metrics that involve true labels cannot be optimized due to lack of any ground truth during training. 
In-processing techniques are also limited to only enforcing demographic parity, which as we showed, remains susceptible to unfairness. One such strategy that has not been applied to OD is 
decoupling, as in \cite{dwork2018decoupled,ustun2019fairness}, where a different detector is trained for each group, while optimizing a joint loss. 

We remark that post-processing and decoupling exhibit treatment disparity as they both assume it to be ethical and legal to use the sensitive attribute at test (decision) time - in particular, to select which threshold or detector to employ on a given new sample. 
When there are differences \textit{among} groups, coming to terms with treatment disparity might be the only get-around to mitigating disparate impact, as argued previously \cite{lipton2018does}.
These solutions, however, do not address unfairness against heterogeneous subpopulations \textit{within} groups, i.e. within-group discrimination. Here, one direction is to explore clustering-based OD algorithms. Alternatively, establishing a more nuanced or granular sensitive attribute, labeling each subpopulation differently.

%% file: appx.tex
\section{Models}
\label{asec:models}

Both the AE and SVDD models use the same architecture, and this architecture is modeled off those in\cite{zhang2021towards}. The architecture is summarized below.

\begin{table}[ht]
\begin{tabular}{@{}lll@{}}
\toprule
\textbf{Part} & \textbf{Layer} & \textbf{Details} \\ \midrule
\multirow{10}{*}{Encoder} 
 & Conv2d       & In: 3, Out: 16, Kernel: 3x3, Stride: 2, \\
 &              & Padding: 1, Bias: False \\
 & ReLU         & In-place: True \\
 & Conv2d       & In: 16, Out: 32, Kernel: 3x3, Stride: 2, \\
 &              & Padding: 1, Bias: False \\
 & BatchNorm2d  & Num Features: 32 \\
 & ReLU         & In-place: True \\
 & Conv2d       & In: 32, Out: 64, Kernel: 3x3, Stride: 2, \\
 &              & Padding: 0, Bias: False \\
 & ReLU         & In-place: True \\
 & Flatten      & Start Dim: 1 \\
 & Linear       & In: 38016, Out: 128, Bias: False \\
 & ReLU         & In-place: True \\
 & Linear       & In: 128, Out: Encoded Space Dim, \\
 &              & Bias: False \\ \midrule
\multirow{13}{*}{Decoder} 
 & Linear             & In: Encoded Space Dim, Out: 128 \\
 & ReLU               & In-place: True \\
 & Linear             & In: 128, Out: 38016 \\
 & ReLU               & In-place: True \\
 & Unflatten          & Dim: 1, Unflattened Size: (64, 22, 27) \\
 & ConvTranspose2d    & In: 64, Out: 32, Kernel: 3x3, Stride: 2, \\
 &                    & Output Padding: 0 \\
 & BatchNorm2d        & Num Features: 32 \\
 & ReLU               & In-place: True \\
 & ConvTranspose2d    & In: 32, Out: 16, Kernel: 3x3, Stride: 2, \\
 &                    & Padding: 1 \\
 & BatchNorm2d        & Num Features: 16 \\
 & ReLU               & In-place: True \\
 & ConvTranspose2d    & In: 16, Out: 3, Kernel: 3x3, Stride: 2, \\
 &                    & Padding: 1, Output Padding: 1 \\
 & Sigmoid            & \\ \bottomrule
\end{tabular}
\end{table}

Datasets are in a random (reset for each initialization) 80-20 split and the model is trained with early stopping if the model does not improve in test loss within three epochs. In practice, the model took, on average 25 minutes to train on a 56-Core 16 GB Tesla P100 GPU.

\section{Raw Data Results}
\label{asec:exp_details}

This subsection of the appendix reports the raw values for DIR and the four properties for each datum, separated by algorithm-dataset interaction. Table \ref{tab:errors} gives the raw sum of squared errors for the individual property models and the entire models used to craft the hypothesis tests.

\begin{table*}
\centering
\caption{Unfairness and property values for CelebA Attributes via Autoencoder}
\label{tab:autoencoder_metrics}
\begin{tabular}{lccccc}
\toprule
Attribute & Unfairness (DIR) & Reconstruction Ratio & SSB & SFV & Label Noise \\
\midrule
5\_o\_Clock\_Shadow & 1.118 & 1.183 & 0.8904 & 0.2252 & 0.4869 \\
Arched\_Eyebrows & 1.124 & 1.0033 & 0.7252 & 0.2287 & 0.4869 \\
Attractive & 1.075 & 1.1356 & 0.5122 & 0.2339 & 0.486 \\
Bags\_Under\_Eyes & 1.308 & 1.077 & 0.799 & 0.2259 & 0.6119 \\
Bald & 1.164 & 1.1017 & 0.9766 & 0.2233 & 0.5019 \\
Bangs & 1.059 & 1.1121 & 0.8518 & 0.2414 & 0.5687 \\
Big\_Lips & 1.219 & 1.1 & 0.7534 & 0.2262 & 0.2721 \\
Big\_Nose & 1.457 & 1.1 & 0.7684 & 0.2247 & 0.4415 \\
Black\_Hair & 1.007 & 1.15 & 0.7568 & 0.2248 & 0.5283 \\
Blond\_Hair & 1.042 & 1.14 & 0.854 & 0.2276 & 0.4273 \\
Blurry & 1.128 & 1.19 & 0.946 & 0.2406 & 1.0181 \\
Brown\_Hair & 1.080 & 1.11 & 0.7962 & 0.224 & 0.6281 \\
Bushy\_Eyebrows & 1.088 & 1.17 & 0.859 & 0.2248 & 0.5107 \\
Chubby & 2.992 & 1.23 & 0.942 & 0.2236 & 0.6076 \\
Double\_Chin & 1.413 & 1.26 & 0.9578 & 0.2244 & 0.709 \\
Eyeglasses & 1.600 & 1.35 & 0.937 & 0.2244 & 0.585 \\
Goatee & 1.479 & 1.27 & 0.9368 & 0.2346 & 0.4938 \\
Gray\_Hair & 1.053 & 1.19 & 0.952 & 0.2247 & 0.5767 \\
Heavy\_Makeup & 1.100 & 1 & 0.6148 & 0.229 & 0.3698 \\
High\_Cheekbones & 1.547 & 1.06 & 0.5536 & 0.235 & 0.6822 \\
Male & 1.117 & 1.01 & 0.5834 & 0.2236 & 0.0211 \\
Mouth\_Slightly\_Open & 1.058 & 1.08 & 0.5222 & 0.2262 & 0.7859 \\
Mustache & 1.280 & 1.3 & 0.9616 & 0.2409 & 0.5055 \\
Narrow\_Eyes & 1.017 & 1.18 & 0.8808 & 0.2252 & 0.7622 \\
No\_Beard & 3.201 & 1.43 & 0.8322 & 0.231 & 0.355 \\
Oval\_Face & 1.672 & 1.07 & 0.7296 & 0.2285 & 0.6119 \\
Pale\_Skin & 1.491 & 1.17 & 0.9586 & 0.2367 & 0.8438 \\
Pointy\_Nose & 1.301 & 1.08 & 0.732 & 0.2272 & 0.5454 \\
Receding\_Hairline & 1.576 & 1.15 & 0.9228 & 0.2248 & 0.6595 \\
Rosy\_Cheeks & 1.035 & 1.19 & 0.9382 & 0.2248 & 0.6718 \\
Sideburns & 1.553 & 1.27 & 0.9396 & 0.2286 & 0.5241 \\
Smiling & 1.317 & 1.07 & 0.5188 & 0.239 & 0.7449 \\
Straight\_Hair & 1.118 & 1.17 & 0.7874 & 0.2268 & 0.6559 \\
Wavy\_Hair & 1.488 & 1.02 & 0.69 & 0.2239 & 0.5728 \\
Wearing\_Earrings & 1.052 & 1.11 & 0.8064 & 0.2269 & 0.6107 \\
Wearing\_Hat & 1.645 & 1.32 & 0.9512 & 0.2269 & 0.7502 \\
Wearing\_Lipstick & 1.213 & 1.07 & 0.5288 & 0.2286 & 0.2678 \\
Wearing\_Necklace & 1.349 & 1.18 & 0.8686 & 0.2246 & 0.6887 \\
Wearing\_Necktie & 1.077 & 1.18 & 0.9244 & 0.2260 & 0.7288 \\
Young & 1.997 & 1.36 & 0.7826 & 0.2250 & 0.164 \\
\bottomrule
\end{tabular}
\end{table*}

\begin{table*}
\centering
\caption{Unfairness and property values for LFW Attributes via Autoencoder}
\label{tab:RawLFWAE}
\resizebox{0.75\textwidth}{!}{
\begin{tabular}{lrrrrrr}\toprule
&Unfairness (DIR) &Reconstruction Ratio &SSB &SFV &Label Noise \\\midrule
Male &1.12200367380267 &1.00535333156585 &0.774632884425169 &0.2031201482 &0.07679999999999998 \\
Asian &1.053645403248 &1.1167961359024 &0.92322909533592 &0.2021178782 & \\
White &1.1264462529671 &1.01018273830413 &0.747926652971163 &0.2088530302 & \\
Black &1.12365634206545 &1.18320667743682 &0.957391767480788 &0.2025963485 &0.08120000000000005 \\
Baby &1.06544960186443 &1.11203300952911 &0.836643079966522 &0.2036614358 &0.09550000000000003 \\
Child &1.12626262626262 &1.12434077262878 &0.898196758730883 &0.2015907168 &0.10470000000000002 \\
Youth &1.1684570024365 &1.09732460975646 &0.786274062238453 &0.2201078296 &0.13770000000000004 \\
Middle Aged &1.05672615298764 &1.10370337963104 &0.866316670470973 &0.2059026539 &0.0645 \\
Senior &1.77747312898089 &1.1931574344635 &0.957467853610286 &0.2064542949 &0.16779999999999995 \\
Black Hair &1.12004451070385 &1.07356524467468 &0.63349311420528 &0.2042071939 & \\
Blond Hair &1.05108769459044 &1.13573598861694 &0.891653351594004 &0.2307599187 & \\
Brown Hair &1.03576168696236 &1.01701772212982 &0.7891653351594 &0.2089367867 & \\
Bald &1.00087648056866 &1.13213109970092 &0.824621471505744 &0.2131045997 &0.11350000000000005 \\
No Eyewear &1.08863610960647 &1.14704251289367 &0.985087118618275 &0.2012821913 &0.06899999999999995 \\
Eyeglasses &1.06348181302805 &1.1471596956253 &0.877729589895762 &0.2019755244 &0.19679999999999997 \\
Sunglasses &1.05997138025237 &1.06211602687835 &0.586700144563646 &0.2035428464 &0.20889999999999997 \\
Mustache &1.0421456164088 &1.03791272640228 &0.581298029369246 &0.2034806907 &0.21950000000000003 \\
Smiling &1.14102186869087 &1.07792913913726 &0.641101727155139 &0.2054580331 &0.2974 \\
Frowning &1.16748745804309 &1.10730016231536 &0.843262573232899 &0.2032266498 &0.07879999999999998 \\
Chubby &1.08701997540087 &1.08353006839752 &0.683557787415354 &0.2049415469 &0.10560000000000003 \\
Blurry &1.04091852227881 &1.10518634319305 &0.811154226584493 &0.2110888839 &0.27580000000000005 \\
Harsh Lighting &1.05681504499685 &1.02375900745391 &0.695579395876131 &0.2157218277 &0.30279999999999996 \\
Soft Lighting &1.05644459380154 &1.03066658973693 &0.598417408506429 &0.2086786151 &0.15849999999999997 \\
Outdoor &1.06132796694575 &1.07666659355163 &0.566613406376017 &0.221262145 &0.22760000000000002 \\
Curly Hair &1.01377517221455 &1.06283998489379 &0.62375408962946 &0.2088014245 &0.14180000000000004 \\
Wavy Hair &1.18200199173129 &1.03124058246612 &0.581830632275736 &0.2119037926 &0.04500000000000004 \\
Straight Hair &1.25206733987405 &1.16164600849151 &0.835806132542037 &0.2045027018 &0.25670000000000004 \\
Receding Hairline &1.132162388614 &1.08063757419586 &0.690329452940728 &0.2026151061 &0.31120000000000003 \\
Bangs &1.16524283964575 &1.00697135925292 &0.672981815415049 &0.2061040878 &0.33009999999999995 \\
Sideburns &1.12084015275504 &1.16936266422271 &0.939739785437114 &0.2015084624 &0.22319999999999995 \\
Fully Visible Forehead &1.21900390887339 &1.1811419725418 &0.858784143650612 &0.2038781226 &0.2298 \\
Partially Visible Forehead &1.05020804838356 &1.04769682884216 &0.536483299094575 &0.2050496221 &0.15200000000000002 \\
Obstructed Forehead &1.197095435684 &1.09454452991485 &0.73674199193487 &0.2027293146 &0.11260000000000003 \\
Bushy Eyebrows &1.21132478772795 &1.02557587623596 &0.645286464277562 &0.2025717795 &0.0998 \\
Arched Eyebrows &1.11604546137808 &1.02139496803283 &0.862360191737046 &0.2056749165 &0.15269999999999995 \\
Narrow Eyes &1.01965937186759 &1.00794005393981 &0.69078596971772 &0.2016550004 &0.19099999999999995 \\
Eyes Open &1.21624935631726 &1.01386857032775 &0.698166324279083 &0.2054687798 &0.2893 \\
Big Nose &1.02608985048702 &1.0604817867279 &0.634406147759263 &0.205928582 &0.06599999999999995 \\
Pointy Nose &1.19868957288718 &1.07568454742431 &0.622764969945978 &0.2045042574 &0.046699999999999964 \\
Big Lips &1.06428433432607 &1.06256353855133 &0.663166704709731 &0.2033233523 &0.08440000000000003 \\
Mouth Closed &1.01175554129597 &1.12742841243743 &0.904511907479266 &0.2021872401 &0.32189999999999996 \\
Mouth Slightly Open &1.06630991503093 &1.04525172710418 &0.571102488016434 &0.2044097185 &0.07479999999999998 \\
Mouth Wide Open &1.01637465524165 &1.01459431648254 &0.714981358898272 &0.2013282001 &0.18810000000000004 \\
Teeth Not Visible &1.09729244959597 &1.1062124967575 &0.759796089172943 &0.2011253536 &0.27669999999999995 \\
No Beard &1.0605139319402 &1.01765537261962 &0.869284029521418 &0.2096437275 &0.2319 \\
Goatee &1.11854311102431 &1.08315765857696 &0.639351746176672 &0.2126889467 &0.04530000000000001 \\
Round Jaw &1.12212437767378 &1.15832161903381 &0.860229780111085 &0.2023314357 &0.050899999999999945 \\
Double Chin &1.01444585996835 &1.04857730865478 &0.519135661568896 &0.2015054762 &0.1965 \\
Wearing Hat &1.31578440808469 &1.13376498222351 &0.950467929696416 &0.2047562778 &0.08360000000000001 \\
Oval Face &1.227980920874 &1.10575580596923 &0.920718253062466 &0.2021733284 &0.12819999999999998 \\
Square Face &1.08180300500834 &1.03950214385986 &0.957087422962793 &0.2091827631 &0.08360000000000001 \\
Round Face &1.00385912356425 &1.03282678127288 &0.504983641482157 &0.2034206629 &0.2126 \\
Color Photo &1.03475440467016 &1.07052874565124 &0.664764513429201 &0.2038946807 &0.10570000000000002 \\
Posed Photo &1.05681639747742 &1.10381340980529 &0.848740774556798 &0.2014933527 &0.15300000000000002 \\
Attractive Man &1.15967929714224 &1.09643280506134 &0.977098075020923 &0.2019033909 &0.35509999999999997 \\
Attractive Woman &1.08906867243748 &1.10497522354125 &0.84105607547744 &0.227733314 &0.13529999999999998 \\
Indian &1.01517435331474 &1.03677427768707 &0.588830556189606 &0.2017122924 &0.14029999999999998 \\
Gray Hair &1.0282016857369 &1.10868871212005 &0.882523016054173 &0.2012593031 &0.18779999999999997 \\
Bags Under Eyes &1.00131664057342 &1.11418402194976 &0.805143422354104 &0.2055422544 &0.13219999999999998 \\
Heavy Makeup &1.14755164575804 &1.11894488334655 &0.879707829262725 &0.2015730679 &0.21609999999999996 \\
Rosy Cheeks &1.02025763283369 &1.05831480026245 &0.507037966978619 &0.2184995234 &0.08520000000000005 \\
Shiny Skin &1.09951096814278 &1.06581234931945 &0.591797915240051 &0.2018399835 &0.10729999999999995 \\
Pale Skin &1.06768325049461 &1.05000007152557 &0.534428973598113 &0.2011277676 &0.1421 \\
5 o' Clock Shadow &1.16855307810665 &1.09448540210723 &0.84432777904588 &0.2016790688 & \\
Strong Nose-Mouth Lines &1.03358017791439 &1.10516810417175 &0.866697101118466 &0.2044023335 & \\
Wearing Lipstick &1.08893014058315 &1.07446813583374 &0.656471125313855 &0.2032339275 & \\
Flushed Face &1.0144694850683 &1.01248931884765 &0.655482005630373 &0.2137254417 & \\
High Cheekbones &1.00916172995591 &1.12738478183746 &0.860229780111085 &0.2017118096 & \\
Brown Eyes &1.08648174717041 &1.01540994644165 &0.636384387126226 &0.2055488884 & \\
Wearing Earrings &1.10247725115406 &1.09833109378814 &0.79563265616678 &0.2028558612 & \\
\bottomrule
\end{tabular}
}
\end{table*}

\begin{table*}\centering
\caption{Unfairness and property values for CelebA Attributes via Deep SVDD}
\label{tab:RawSVDDCelebA}
\begin{tabular}{lrrrrrr}\toprule
&SVDD &Reconstruction &SSB &Spurious Feature Variance &Label Noise \\\midrule
5\_o\_Clock\_Shadow &1.68123553498308 &1.46047670114505 &0.8904 &0.1409505606 &0.4869 \\
Arched\_Eyebrows &1.25764192139738 &1.29294249928091 &0.7252 &0.1452494413 &0.4869 \\
Attractive &1.09368792760979 &1.05381571022971 &0.5122 &0.1520317346 &0.486 \\
Bags\_Under\_Eyes &1.06134410518395 &1.12471149407601 &0.798999999999999 &0.1400723457 &0.6119 \\
Bald &2.352 &1.02880658436214 &0.9766 &0.1451713741 &0.5019 \\
Bangs &1.39449541284403 &1.32236633976589 &0.8518 &0.1576949656 &0.5687 \\
Big\_Lips &1.70156624102154 &1.08017998183669 &0.7534 &0.1442556977 &0.2721 \\
Big\_Nose &1.30569948186528 &1.16960464068483 &0.7684 &0.1389202923 &0.4415 \\
Black\_Hair &1.00635593220339 &1.10986682808716 &0.7568 &0.1428498179 &0.5283 \\
Blond\_Hair &1.03992089562244 &1.29937377627469 &0.854 &0.1420869976 &0.4273 \\
Blurry &1.35800508259212 &1.12928843710292 &0.946 &0.1826313585 &1.0181 \\
Brown\_Hair &1.07992104600792 &1.00426740416926 &0.7962 &0.1399643421 &0.6281 \\
Bushy\_Eyebrows &1.26066424494032 &1.1293009118541 &0.859 &0.1396305859 &0.5106999999999999 \\
Chubby &1.15950659293917 &1.0393457117595 &0.942 &0.143850103 &0.6075999999999999 \\
Double\_Chin &1.46185598532334 &1.24815246204514 &0.9578 &0.1393095106 &0.7090000000000001 \\
Eyeglasses &1.4847619047619 &1.13053239255933 &0.937 &0.1726125926 &0.585 \\
Goatee &1.30087633885102 &1.29235531479741 &0.9368 &0.148198694 &0.4938 \\
Gray\_Hair &2.44949494949494 &1.63565217391304 &0.952 &0.1400457323 &0.5767 \\
Heavy\_Makeup &1.37794331165961 &1.21201795786807 &0.6148 &0.1471818388 &0.3698 \\
High\_Cheekbones &1.41521739130434 &1.07322226737098 &0.5536 &0.148946777 &0.6821999999999999 \\
Male &1.16378620579292 &1.12330668559143 &0.583399999999999 &0.1411117315 &0.021100000000000008 \\
Mouth\_Slightly\_Open &1.47328992862486 &1.01889931435045 &0.5222 &0.1415492892 &0.7859 \\
Mustache &1.28 &1.04602510460251 &0.9616 &0.1618342251 &0.5055000000000001 \\
Narrow\_Eyes &1.3557779799818 &1.08768131630222 &0.8808 &0.1405434906 &0.7622 \\
No\_Beard &1.26765068774848 &1.32170279829207 &0.8322 &0.1428951621 &0.355 \\
Oval\_Face &1.00961538461538 &1.05142857142857 &0.7296 &0.1448870301 &0.6119 \\
Pale\_Skin &1.49135109864422 &1.06838387528924 &0.9586 &0.1604245007 &0.8438 \\
Pointy\_Nose &1.28422782037239 &1.26457127210139 &0.732 &0.1431550533 &0.5454 \\
Receding\_Hairline &1.10142050741269 &1.33176813471502 &0.9228 &0.1404222101 &0.6595 \\
Rosy\_Cheeks &1.153123680878 &1.25196285352469 &0.9382 &0.1406327337 &0.6718 \\
Sideburns &1.36525725929699 &1.50509087726463 &0.9396 &0.1386207491 &0.5241 \\
Smiling &1.11647331786542 &1.01603413341645 &0.518799999999999 &0.1510140896 &0.7449 \\
Straight\_Hair &1.13916759320035 &1.16279926135717 &0.7874 &0.1428056359 &0.6558999999999999 \\
Wavy\_Hair &1.6726155889433 &1.30170504067402 &0.69 &0.1401683241 &0.5728 \\
Wearing\_Earrings &1.08250497017892 &1.01847107438016 &0.8064 &0.1419264823 &0.6107 \\
Wearing\_Hat &5.03622577927548 &1.54812552653748 &0.9512 &0.2158842981 &0.7502 \\
Wearing\_Lipstick &1.26436951774677 &1.16687742370595 &0.528799999999999 &0.1471352577 &0.26780000000000004 \\
Wearing\_Necklace &1.00260846420015 &1.07914052831476 &0.8686 &0.1395401657 &0.6887 \\
Wearing\_Necktie &1.52579365079365 &1.36231575963718 &0.9244 &0.1407860667 &0.7288 \\
Young &1.0892026578073 &1.17668546526531 &0.7826 &0.1402778327 &0.16400000000000003 \\
\bottomrule
\end{tabular}
\end{table*}

\begin{table*}\centering
\caption{Unfairness and property values for LFW Attributes via Deep SVDD}
\label{tab:RawSVDDLFW}
\resizebox{0.72\textwidth}{!}{
\begin{tabular}{lrrrrrr}\toprule
&DIR &Incompressibility &SSB &SFV &Label Noise \\\midrule
Male &1.17931562745317 &1.0924447774887 &0.774632884425169 &0.1429237619 &0.07679999999999998 \\
Asian &1.23055692048871 &1.01016497611999 &0.92322909533592 &0.1415031001 & \\
White &1.00406917599186 &1.08468961715698 &0.747926652971163 &0.1511053567 & \\
Black &1.34819532908704 &1.03384220600128 &0.957391767480788 &0.1421764963 &0.08120000000000005 \\
Baby &1.06271364829537 &1.03743159770965 &0.836643079966522 &0.143683949 &0.09550000000000003 \\
Child &1.13670569529881 &1.03432464599609 &0.898196758730883 &0.140765438 &0.10470000000000002 \\
Youth &1.05082822021653 &1.03086674213409 &0.786274062238453 &0.1678981342 &0.13770000000000004 \\
Middle Aged &1.10867550207333 &1.02771830558776 &0.866316670470973 &0.1468662707 &0.0645 \\
Senior &1.00186866902908 &1.04160547256469 &0.957467853610286 &0.1476565742 &0.16779999999999995 \\
Black Hair &1.02750194844192 &1.04895257949829 &0.63349311420528 &0.1444525348 & \\
Blond Hair &1.08838038386602 &1.01417303085327 &0.891653351594004 &0.184528688 & \\
Brown Hair &1.03209559606518 &1.09958708286285 &0.7891653351594 &0.1512480983 & \\
Bald &1.00790551940226 &1.00073754787445 &0.824621471505744 &0.1573695429 &0.11350000000000005 \\
No Eyewear &1.40606623336428 &1.00252616405487 &0.985087118618275 &0.1403350747 &0.06899999999999995 \\
Eyeglasses &1.43913177607322 &1.00310981273651 &0.877729589895762 &0.1413070618 &0.19679999999999997 \\
Sunglasses &1.12142575468585 &1.05022633075714 &0.586700144563646 &0.1435135175 &0.20889999999999997 \\
Mustache &1.11839255634876 &1.07019913196563 &0.581298029369246 &0.1434256518 &0.21950000000000003 \\
Smiling &1.07582623948232 &1.03398072719573 &0.641101727155139 &0.1462316354 &0.2974 \\
Frowning &1.38021050679278 &1.04254591464996 &0.843262573232899 &0.14306304 &0.07879999999999998 \\
Chubby &1.37894686222649 &1.04961144924163 &0.683557787415354 &0.1454682116 &0.10560000000000003 \\
Blurry &1.07484216395665 &1.01648831367492 &0.811154226584493 &0.1543799572 &0.27580000000000005 \\
Harsh Lighting &1.76541734255385 &1.06726253032684 &0.695579395876131 &0.1612156139 &0.30279999999999996 \\
Soft Lighting &1.15505859850802 &1.0463809967041 & &0.1456943556 &0.1642 \\
Outdoor &1.20495433082845 &1.06911957263946 &0.598417408506429 &0.1508525053 &0.15849999999999997 \\
Curly Hair &1.13923719958202 &1.05589497089385 &0.566613406376017 &0.1696171921 &0.22760000000000002 \\
Wavy Hair &1.06940992787003 &1.04987812042236 &0.62375408962946 &0.1510261253 &0.14180000000000004 \\
Straight Hair &1.04934265833276 &1.07179963588714 &0.581830632275736 &0.1555814983 &0.04500000000000004 \\
Receding Hairline &1.17698276832539 &1.00636541843414 &0.835806132542037 &0.1448722679 &0.25670000000000004 \\
Bangs &1.09157918248827 &1.06093919277191 &0.690329452940728 &0.1422119786 &0.31120000000000003 \\
Sideburns &1.15947653456037 &1.08820021152496 &0.672981815415049 &0.1471594972 &0.33009999999999995 \\
Fully Visible Forehead &1.55668147556531 &1.00061905384063 &0.939739785437114 &0.1406534992 &0.22319999999999995 \\
Partially Visible Forehead &1.25747607655502 &1.03077602386474 &0.858784143650612 &0.1439849571 &0.2298 \\
Obstructed Forehead &1.11325281649095 &1.05480468273162 &0.536483299094575 &0.1456306697 &0.15200000000000002 \\
Bushy Eyebrows &1.00786702803827 &1.03234314918518 &0.73674199193487 &0.1423631794 &0.11260000000000003 \\
Arched Eyebrows &1.06208761023718 &1.07342624664306 &0.645286464277562 &0.1421420989 &0.0998 \\
Narrow Eyes &1.08786442753544 &1.09905493259429 &0.862360191737046 &0.1465277227 &0.15269999999999995 \\
Eyes Open &1.223370100546 &1.07901871204376 &0.69078596971772 &0.1408611888 &0.19099999999999995 \\
Big Nose &1.03111518672274 &1.0845707654953 &0.698166324279083 &0.1462382611 &0.2893 \\
Pointy Nose &1.11446611115883 &1.05301141738891 &0.634406147759263 &0.1469018736 &0.06599999999999995 \\
Big Lips &1.14029929024963 &1.05008065700531 &0.622764969945978 &0.1448668399 &0.046699999999999964 \\
Mouth Closed &1.0816224959562 &1.06419742107391 &0.663166704709731 &0.1431995614 &0.08440000000000003 \\
Mouth Slightly Open &1.01337628971086 &1.03440833091735 &0.904511907479266 &0.1416061479 &0.32189999999999996 \\
Mouth Wide Open &1.03990024937655 &1.06561398506164 &0.571102488016434 &0.1447347991 &0.07479999999999998 \\
Teeth Not Visible &1.04770316767762 &1.07595670223236 &0.714981358898272 &0.1403963187 &0.18810000000000004 \\
No Beard &1.0876431987543 &1.03276085853576 &0.759796089172943 &0.1401160741 &0.27669999999999995 \\
Goatee &1.19802672343941 &1.09399461746215 &0.869284029521418 &0.1522871416 &0.2319 \\
Round Jaw &1.06897059287373 &1.04447555541992 &0.639351746176672 &0.156724269 &0.04530000000000001 \\
Double Chin &1.02390223246378 &1.00809562206268 &0.860229780111085 &0.141805179 &0.050899999999999945 \\
Wearing Hat &1.08353184055899 &1.06220078468322 &0.519135661568896 &0.1406492755 &0.1965 \\
Oval Face &1.12880495352612 &1.02028930187225 &0.950467929696416 &0.1452499895 &0.08360000000000001 \\
Square Face &1.03973957569458 &1.0401998758316 &0.920718253062466 &0.141584407 &0.12819999999999998 \\
Round Face &1.1325489572568 &1.10276663303375 &0.957087422962793 &0.1514307107 &0.08360000000000001 \\
Color Photo &1.13798432728612 &1.07599568367004 &0.504983641482157 &0.143338242 &0.2126 \\
Posed Photo &1.0394726007875 &1.0795783996582 &0.664764513429201 &0.1440141143 &0.10570000000000002 \\
Attractive Man &1.08110687391574 &1.03554499149322 &0.848740774556798 &0.1406362299 &0.15300000000000002 \\
Attractive Woman &1.51685778921912 &1.05630433559417 &0.977098075020923 &0.1412037472 &0.35509999999999997 \\
Indian &1.04062717938913 &1.05820059776306 &0.84105607547744 &0.1797280974 &0.13529999999999998 \\
Gray Hair &1.1973171397336 &1.06187999248504 &0.588830556189606 &0.1409442876 &0.14029999999999998 \\
Bags Under Eyes &1.45747314192372 &1.02336192131042 &0.882523016054173 &0.140306542 &0.18779999999999997 \\
Heavy Makeup &1.00048536793256 &1.02952170372009 &0.805143422354104 &0.1463585953 &0.13219999999999998 \\
Rosy Cheeks &1.16560850348722 &1.03843355178833 &0.879707829262725 &0.1407405867 &0.21609999999999996 \\
Shiny Skin &1.26315502068762 &1.05610179901123 &0.507037966978619 &0.1653680619 &0.08520000000000005 \\
Pale Skin &1.18351263647553 &1.0652779340744 &0.591797915240051 &0.1411130869 &0.10729999999999995 \\
5 o' Clock Shadow &1.17272883141058 &1.05673563480377 &0.534428973598113 &0.1401202399 &0.1421 \\
Strong Nose-Mouth Lines &1.22455279703978 &1.03042745590209 &0.84432777904588 &0.1408965065 & \\
Wearing Lipstick &1.19464912714504 &1.04016602039337 &0.866697101118466 &0.1447695088 & \\
Flushed Face &1.09233277829563 &1.04554724693298 &0.656471125313855 &0.1430755171 & \\
High Cheekbones &1.11059337257828 &1.08112835884094 &0.655482005630373 &0.1582355048 & \\
Brown Eyes &1.17224142515288 &1.01272892951965 &0.860229780111085 &0.1409409852 & \\
Wearing Earrings &1.16634746922024 &1.0849984884262 &0.636384387126226 &0.1463791189 & \\
\bottomrule
\end{tabular}
}
\end{table*}

\label{asec:cluster_details}

\begin{table*}
\centering
\caption{Squared Error (SE) For Various Properties \& The Proposed Whole Model}
\label{tab:errors}
\begin{tabular}{lrrrrrr}\toprule
&Incompressibility &SSB &SFV &Label Noise &Whole Model \\\midrule
5\_o\_Clock\_Shadow &0.0684682 &0.0459237 &0.0002894 &3e-7 &3e-7 \\
Arched\_Eyebrows &0.0126105 &0.0538039 &0.0131643 &0.0106393 &0.0106393 \\
Attractive &0.0004621 &0.0338707 &0.0483136 &0.0275897 &0.0004621 \\
Bags\_Under\_Eyes &0.0202342 &0.0623039 &0.0050525 &0.0093071 &0.0050525 \\
Bald &0.4319569 &0.00024 &0.0943209 &0.0748498 &0.00024 \\
Bangs &0.0006496 &0.1283069 &0.0774581 &0.0298166 &0.0006496 \\
Big\_Lips &0.1088419 &0.0091254 &0.0032081 &0.0052196 &0.0032081 \\
Big\_Nose &0.2114748 &0.0001715 &0.0252706 &0.020549 &0.0001715 \\
Black\_Hair &0.0142123 &0.0593484 &0.0061956 &0.0106132 &0.0061956 \\
Blond\_Hair &0.0100063 &0.0946047 &0.0130189 &0.0110102 &0.0100063 \\
Blurry &0.5419093 &0.0074364 &0.0490523 &0.0729299 &0.0074364 \\
Brown\_Hair &0.0001064 &0.1154422 &0.0218438 &0.0360707 &0.0001064 \\
Bushy\_Eyebrows &0.0213848 &0.0773384 &0.0035219 &0.006768 &0.0035219 \\
Chubby &0.0013206 &0.1581314 &0.015918 &0.0287418 &0.0013206 \\
Double\_Chin &0.0035046 &0.1521108 &0.0137349 &0.0272636 &0.0035046 \\
Eyeglasses &0.0113092 &0.120019 &0.0071329 &0.0140104 &0.0071329 \\
Goatee &0.1604753 &0.0201623 &0.0013985 &0.0092697 &0.0013985 \\
Gray\_Hair &0.0002775 &0.1933109 &0.0296317 &0.0411467 &0.0002775 \\
Heavy\_Makeup &0.0000469 &0.0613383 &0.0363665 &0.0266154 &0.0000469 \\
High\_Cheekbones &0.0930655 &0.0000754 &0.0010459 &0.0000749 &0.0000749 \\
Male &0.0336658 &0.0118035 &0.0005703 &0.0000012 &0.0000012 \\
Mouth\_Slightly\_Open &0.0443864 &0.0033491 &0.0006975 &0.0047076 &0.0006975 \\
Mustache &0.282736 &0.0040766 &0.0060553 &0.0346049 &0.0040766 \\
Narrow\_Eyes &0.2105005 &0.005241 &0.0237357 &0.0113154 &0.005241 \\
No\_Beard &0.6583177 &0.03911 &0.1239485 &0.1577867 &0.03911 \\
Oval\_Face &0.3213105 &0.0066273 &0.0411204 &0.0393565 &0.0066273 \\
Pale\_Skin &0.2731508 &0.0047057 &0.0111297 &0.0200174 &0.0047057 \\
Pointy\_Nose &0.0001962 &0.0922673 &0.0293014 &0.0316888 &0.0001962 \\
Receding\_Hairline &0.5347036 &0.0090082 &0.1198915 &0.0943515 &0.0090082 \\
Rosy\_Cheeks &0.0309238 &0.0892779 &0.0015001 &0.0063849 &0.0015001 \\
Sideburns &0.1497514 &0.023381 &0.005897 &0.0069555 &0.005897 \\
Smiling &0.1429648 &0.0036202 &0.0001827 &0.0027342 &0.0001827 \\
Straight\_Hair &0.2121866 &0.0005044 &0.0203373 &0.0143183 &0.0005044 \\
Wavy\_Hair &0.0181157 &0.0392565 &0.0036124 &0.0094807 &0.0036124 \\
Wearing\_Earrings &0.0000825 &0.1285602 &0.0341021 &0.0405731 &0.0000825 \\
Wearing\_Hat &0.0069229 &0.1368304 &0.0145395 &0.0234406 &0.0069229 \\
Wearing\_Lipstick &0.0278786 &0.0084025 &0.0057513 &0.0016339 &0.0016339 \\
Wearing\_Necklace &0.0691438 &0.0408226 &0.0005308 &0.0004549 &0.0004549 \\
Wearing\_Necktie &0.1449726 &0.0222941 &0.0086847 &0.00313 &0.00313 \\
Young &0.1940491 &0.0011516 &0.0204862 &0.0260069 &0.0011516 \\
\bottomrule
\end{tabular}
\end{table*}